\pdfoutput=1

\documentclass[11pt]{article}

\usepackage[review]{acl}

\usepackage{times}
\usepackage{latexsym}

\usepackage{hyperref}

\usepackage[T1]{fontenc}

\usepackage[utf8]{inputenc}

\usepackage{microtype}
\usepackage{amsmath}

\usepackage{inconsolata}

\title{Rubric-Specific Approach to Automated Essay Scoring with Augmentation Training}

\author{Brian Cho, Youngbin Jang, Jaewoong Yoon \\
  Databank, Inc. \\
  \texttt{\{brianhscho, ybjang, yjwoong97\}@data-bank.ai} \\}

\begin{document}
\maketitle
\begin{abstract}
Neural based approaches to automatic evaluation of subjective responses have shown superior performance and efficiency compared to traditional rule-based and feature engineering oriented solutions. However, it remains unclear whether the suggested neural solutions are sufficient replacements of human raters as we find recent works do not properly account for rubric items that are essential for automated essay scoring during model training and validation. In this paper, we propose a series of data augmentation operations that train and test an automated scoring model to learn features and functions overlooked by previous works while still achieving state-of-the-art performance in the Automated Student Assessment Prize dataset.
\end{abstract}

\section{Introduction}

Automated Essay Scoring (AES) is defined as the task of applying automation algorithms to evaluate the quality of written essay responses without the intervention of a human grader. Recently, neural applications involving deep neural networks and representation learning quickly proved to be flexible and effective \cite{ramesh2021automated} in AES. Consequently, series of neural based approaches to AES including recurrent neural networks \cite{taghipour-ng-2016-neural}, attention mechanism \cite{dong-etal-2017-attention}, and pre-trained language models \cite{yang-etal-2020-enhancing, jeon-strube-2021-countering} have been researched and tested on the Automated Student Assessment Prize (ASAP) dataset.\footnote{\url{https://www.kaggle.com/c/asap-aes}}

However, the ongoing AES performance competition overlooks several critical problems. Specifically, multiple attributes commonly found from scoring rubrics are left out from consideration during AES model training and validation. Instead of evaluating the AES model's alignment with items outlined in the rubric, previous neural approaches to ASAP focus on achieving state-of-the-art similarity scores between a human rater and an AES model. While similarity score is one important aspect of functioning AES systems, it alone does not guarantee that an AES model can replace a human rater \cite{bennett1997validity, attali2007construct, zhang2013contrasting, perelman2014state, madnani2018automated}. Therefore, previous neural approaches must be evaluated for additional AES functions other than similarity \cite{kabra2022evaluation} before being deployed for service.

A deeper investigation into previous works reveals a potential source for the aforementioned problem. One common trait shared by previous neural approaches to ASAP is the implementation of prompt-specific models \cite{attali2005automated, ridley2021automated}. The approach is defined by how the training dataset is segmented. Given a dataset comprised of essays to \(n\) question prompts, prompt-specific approach segments the dataset into \(n\) subsets based on prompt (prompt-segmented dataset) and trains one AES model on each subset, resulting in \(n\) prompt-specific models for \(n\) question prompts even when the prompts share the same rubric. The \(n\) models train to learn the same scoring rubric, but the scoring standard learned by each model will likely diverge as the model optimizes on each data segments \cite{attali2010performance, chen2013automated}, resulting in models that no longer embody the original scoring rubric. Moreover, segmenting the dataset based on features eliminates the need for AES models to account for those features during training and validation. For instance, once the dataset is segmented based on question prompts, the AES model is never tested on its ability to assess relevance of essay responses in relation to varying question prompts. Similarly, once the training dataset is segmented based on features relating to a specific rubric item, the resulting AES model will not be able to account for the rubric item during training and inference \cite{madnani2018automated}.

In this research, we propose an alternative approach to AES termed rubric-specific model. Rubric-specific models are trained and tested on datasets segmented by scoring rubrics (rubric-segmented dataset), resulting in \(n\) rubric-specific models for \(n\) scoring rubrics. Each rubric-specific model is trained to be the best and only representation of it's respective scoring rubric regardless of how many prompts are tied to the same rubric. The proposed approach is general and efficient as it is aligned with human raters who are trained to learn each scoring rubric instead of each question prompt. Most importantly, since rubric-segmented datasets include features precluded in prompt-segmented datasets, rubric-specific models must consider the following rubric items overlooked by prompt-specific models:
\begin{itemize}
    \item Rubric-segmented dataset includes essay responses to multiple question prompts. Therefore, rubric-specific models must be able to distinguish various response-prompt combinations and assess the relevance of an essay in relation to the question prompt. Relevance assessment is not only essential in essay scoring, but is also crucial in AES service application. For instance, the inability to detect and evaluate irrelevant responses leaves the AES model unprepared against adversarial attacks and could potentially debunk the effectiveness and reliability of AES systems in their entirety \cite{ding-etal-2020-dont, kabra2022evaluation}.
    \item Rubric-segmented dataset includes essay responses written by students from varying grade levels. Along with individual writing skills, student grade level is also a significant predictor of essay scores \cite{burdick2013measuring}. Therefore, the quality of writing a human rater expects from a well-written essay should be different and adjusted based on the student's grade level. Similar to a human rater, a rubric-specific model must be able to identify and incorporate grade level differences in automated scoring \cite{zhang2013contrasting}.
\end{itemize}
In addition to the previously precluded factors, our research seeks to address another rubric item that is fundamental to essay scoring, yet insufficiently investigated during the performance competition at ASAP:
\begin{itemize}
    \item Human raters are expected to detect and penalize incoherently ordered words or sentences. However, the same cannot be expected from neural AES systems \cite{pham2020out}. Consequently, rubric-specific models must be equipped with and tested on the ability to penalize permuted text and distinguish adversarial inputs \cite{farag-etal-2018-neural, ding-etal-2020-dont, singla2021aes, kabra2022evaluation}.
\end{itemize}
Our experiment demonstrates training an AES model to learn the above rubric items while maintaining significant similarity score requires more than simply training on a rubric-segmented dataset. We introduce three data augmentation methods, \(Prompt\ Swap\), \(Grade\ Match\), and \(Response\ Distortion\), to guide the AES model to learn the intended features without suffering from robustness-accuracy trade-off \cite{su2018robustness}. Moreover, we introduce a neural network architecture, \(Response-Prompt\ AES\), capable of processing the suggested augmentation training. Our experiment results show the proposed augmentation methods resolve the functional limitations of previous neural approaches to AES. In addition to the added functions, we also demonstrate our proposed AES model achieves state-of-the art performance in the ASAP dataset.
\section{Related Work} \label{related_work}

\paragraph{Neural AES} Neural approaches to AES adopted learned representations such as pre-trained word vectors \cite{taghipour-ng-2016-neural, mathias-etal-2020-happy} and contextual embeddings from pre-trained language models \cite{yang-etal-2020-enhancing, jeon-strube-2021-countering, 9530411} to replace conventional handcrafted features utilized in AES. In addition to learned features, recent works experimented with training strategies such as multi-task learning \cite{muangkammuen-fukumoto-2020-multi, yang-etal-2020-enhancing, mathias-etal-2020-happy} to achieve enhanced performance in the ASAP dataset. 

\paragraph{Generic AES} While neural applications in AES proved to be effective, prompt-specific AES models required large amounts of labeled training data and were limited to scoring essays from only one question prompt. To address the problems of efficiency, researches including \citet{jin-etal-2018-tdnn} and \citet{ridley2021automated} proposed a prompt-independent approach to AES utilizing essay responses from multiple prompts for AES model training. However, while generic AES model training involved essay responses to multiple question prompts, the topic of irrelevant responses or adversarial inputs was never properly discussed.

\paragraph{Robust AES} The performance race at ASAP sparked another important discussion in AES. Multiple researches raised questions regarding the robustness of neural AES systems. According to related works, state-of-the-art AES models were unable to detect essays with randomly shuffled word ordering \cite{farag-etal-2018-neural, ding-etal-2020-dont}, off-topic content \cite{liu2019automated, kabra2022evaluation}, and abnormal inputs \cite{perelman2014state}. While related works proposed various adversarial training methods as potential remedies, increase in AES model robustness was accompanied with loss in accuracy, architecture overhead, and added optimization tasks \cite{liu2019automated, ding-etal-2020-dont, sun2022improving}.
\section{Response-Prompt AES}\label{sec:model}

We first introduce the neural network architecture implemented in our experiments and describe the computation flow and reasoning behind the model structure. Our AES model includes a pre-trained language model, a response self-attention layer, and a response-prompt attention layer which are all fine-tuned on the ASAP dataset. Implementation details on each module are listed below in order of computation.

\paragraph{Pre-trained Language Model} To generate contextual embeddings from essays and prompts, we utilize the widely successful pre-trained language model BERT \cite{Devlin2019-sf} and its implementation (bert-base-uncased) in the Python language.\footnote{\url{https://github.com/huggingface/transformers}} Essay responses and question prompts are tokenized into list of tokens and used as inputs to BERT. To address the maximum token length restriction imposed on BERT, token lists longer than 512 are segmented and stacked into token groups of size 512. After forward passing through BERT, we collapse the segment axis of the output embedding matrix.

\paragraph{Response Self-Attention Layer} The collapsed embedding matrix passes through a custom designed self-attention layer without predefined length restriction. The response self-attention layer implements self-attention \cite{vaswani2017attention} with relative position embeddings \cite{shaw-etal-2018-self} to simulate human reading pattern on long texts with multiple sentences and paragraphs. Response self-attention is computed as follows:

\begin{equation} \label{eq:1}
e_{ij} = \frac{x_iW^Q(x_jW^K)^T + x_iW^Q(R^K_{ij})^T}{\sqrt{d}}
\end{equation}
\begin{equation} \label{eq:2}
a_{ij} = \frac{\exp{e_{ij}}}{\sum_{k=1}^{n}\exp{e_{ik}}}
\end{equation}
\begin{equation} \label{eq:3}
z_{i} = {\sum_{j=1}^{n}a_{ij}(x_jW^V + R^V_{ij})}
\end{equation}
where \(\{x_1, x_2, ..., x_n\}\) is the embedding matrix output from BERT, \(W^Q, W^K, W^V\) trainable parameters from the attention layer, and \(R^K\) and \(R^V\) relative position representation matrices also trained during the fine-tuning process.

Finally, the contextualized embedding vector from equation \ref{eq:3} corresponding to the CLS token position index, \(z_1\), is used as essay response representation.

\paragraph{Response-Prompt Attention Layer} Attention mechanism \cite{DBLP:journals/corr/BahdanauCB14} is implemented in the response-prompt attention layer. Essay response representation matrix \(\{z_1, z_2, ..., z_n\}\) attends the prompt embedding matrix \(P\) to compute response-prompt attention vector as shown below. 

\begin{equation} \label{eq:4}
e_{ij} = \frac{z_iW^Q(P_jW^K)^T}{\sqrt{d}}
\end{equation}
\begin{equation} \label{eq:5}
a_{ij} = \frac{\exp{e_{ij}}}{\sum_{k=1}^{n}\exp{e_{ik}}}
\end{equation}
\begin{equation} \label{eq:6}
r_i = {\sum_{j=1}^{n}a_{ij}P_jW^V}
\end{equation}
The resulting response-prompt attention vector corresponding to the CLS token position index, \(r_1\), is used as response-prompt relevance representation.

\paragraph{Regression Layer} Lastly, the essay response representation vector \(z_1\) is concatenated with the response-prompt attention vector \(r_1\) to form the final representation vector used for score prediction. The concatenated vector passes through a dense layer to compute the model output, \(\hat{o}\), as shown in equation \ref{eq:7}.

\begin{equation} \label{eq:7}
\hat{o} = concat(z_1, r_1)W + b
\end{equation}

The Response-Prompt AES model is trained with the Mean Squared Error (MSE) loss function. Specifically, given training label score \(o_i\), the model is trained to minimize the following loss function over the training dataset:

\begin{equation} \label{eq:8}
MSE = \frac{1}{n}\sum_{i=1}^{n}(o_i-\hat{o}_i)^2
\end{equation}
\section{Experiment}

\label{experiment objective}In this section, we outline the details of training a rubric-specific model. Specifically, we describe our experimental procedure for implementing Prompt Swap, Grade Match, and Response Distortion on the Response-Prompt AES model. Our experiment is focused on the following investigations:
\begin{itemize}
    \item Measure the isolated effect of each data augmentation method and establish it's functional significance in AES.
    \item Assess the isolated and combined effect of data augmentation methods on the ASAP dataset against benchmark performance.
\end{itemize}
Hyper-parameter settings for all training and testing experiments are summarized in Appendix \ref{appendix:appendix_B}. Our source code and experiment logs are publicly available for review and replication.\footnote{Anonymized URL}

\subsection{Dataset} Our experiment uses the Automated Student Assessment Prize dataset from Kaggle. This dataset includes essay responses to eight different question prompts, and each essay response is labeled with an evaluation score given by a human grader according to the prompt's respective scoring rubric. Statistical summary and metadata of the dataset are provided in Tables \ref{Appendix A.1} and \ref{Appendix A.2} of Appendix \ref{appendix:appendix_A}, respectively.

Aligned with the definition of rubric-specific models, we segment the dataset into six subsets corresponding to the number of unique scoring rubrics and train one AES model from each subset. For easier comparison, our six AES models are labeled Prompt 1, 2, 7, 8, 3-4, and 5-6 model.

\subsection{Performance Evaluation} For performance assessment in the ASAP dataset, we use 5-fold cross validation to evaluate our AES model with 60\% / 20\% / 20\% data split amongst training, develop, and test sets. Fold indices are adopted from \citet{taghipour-ng-2016-neural}. All of the selected performance benchmarks listed in the \hyperref[results-and-analysis]{Results and Analysis} section follow the same fold indices for accurate performance comparison. Consistent with previous works, we select our best AES model based on the performance in the develop set and adopt \textbf{Quadratic Weighted Kappa} (QWK) \footnote{\url{https://www.kaggle.com/competitions/asap-aes/overview/evaluation}} as evaluation metric.

For performance assessment of each data augmentation method, we employ specific metrics further explained in the following subsections. Data augmentation performances are also reported in averages computed over 5 folds.

\subsection{Data Augmentation Implementation}

\subsubsection{Prompt Swap} 
The fundamental fact that an essay's score is dependent on the question prompt is often overlooked. For example, an essay response to question prompt 3 that received a perfect score is no longer considered relevant when paired with question prompt 4 regardless of writing quality. The idea of relevance is also embedded in the ASAP scoring rubric which is shown in Table \ref{Appendix A.3} of Appendix \ref{appendix:appendix_A}. While distinguishing essays to prompt 3 from essays to prompt 4 is easy task for human raters, the same cannot be expected from AES systems. Accordingly, we apply Prompt Swap to Prompt 3-4 and 5-6 models to train relevance aware AES models.

Prompt Swap generates prompt mismatched essay samples with known labels for augmentation training. For a given training batch \(b=\{t_1, t_2, ..., t_n\}\) where \(t_i=\{essay, prompt, score\}\), we select \(k\) samples from the training batch, swap the prompt to mismatch the essay response, and add the generated irrelevant response-prompt sample to the training batch with known label of score zero (lowest possible score). For example, if \(s=\{essay_4, prompt_4, score=3\}\) is a relevant response-prompt sample addressing prompt 4 with a perfect score, the AES model should also be able to train from and accurately predict irrelevant response-prompt sample such as \({s^\prime}=\{essay_4, prompt_3, score=0\}\). When selecting the \(k\) samples for augmentation, Prompt Swap is only applied to essay responses with original label scores greater than the average score. Such condition is necessary as low score essays have low writing quality regardless of relevance, making it difficult to isolate the effect of Prompt Swap.

The contribution of Prompt Swap is reported in two folds. First, \textbf{irrelevant response detection rate} is measured with prompt swapped samples generated from the test set. The AES model should predict the lowest label score for the prompt swapped samples, which we count as detection success. All other score predictions are counted as detection failures. Irrelevant response detection rate is computed and compared against baseline models trained without Prompt Swap. Second, we investigate whether Prompt Swap improves both robustness and accuracy of AES models by comparing test set QWK against baseline models trained without Prompt Swap.

\subsubsection{Grade Match} Unlike Prompts 3 and 4 which are written by students in the same grade level, Prompts 5 and 6 are written by students in different grade levels as shown in Table \ref{Appendix A.2} of Appendix \ref{appendix:appendix_A}. Therefore, we hypothesized that while essay responses to prompts 5 and 6 are graded with the same scoring rubric, a human rater must adjust the expectation for a well-written essay based on the student's grade level. Accordingly, we apply Grade Match in Prompt 5-6 model to not only differentiate scores, but also differentiate essays from different grade levels.

The process of recognizing differences between essays is analogous to training an AES model to learn distances between essay representations in the  embedding space. Particularly, Grade Match seeks to map essays from the same grade level close together while distancing them from essays from other grade levels. Grade Match is inspired by the methodologies proposed in Supervised Contrastive Learning \cite{NEURIPS2020_d89a66c7}, which leverages label information to contrast batch items from one class against batch items from another class. Following the same strategy, Prompt 5-6 model utilizes score and grade level as labels during Grade Match.

Given essay batch \(E=\{e_1, e_2, ..., e_n\}\), corresponding score label \(S=\{s_1, s_2, ..., s_n\}\), corresponding grade level label \(G=\{g_1, g_2, ..., g_n\}\), and augmentation sample count \(k\), the essay response representation set \(Z=\{z_1, z_2, ..., z_n\}\) is calculated for each corresponding batch item in \(E\) according to Equation \ref{eq:3}. Next, cosine similarity \(c_s\) is calculated for batch items with the same score and same grade level as follows.

\begin{equation} \label{eq:9}
c_s = {\sum_{\substack{g_i=g_j \in G, \\ i \neq j}}^{}\sum_{\substack{s_i=s_j \in S, \\ i \neq j}}Cos(z_i, z_j)}
\end{equation}
Similarly, cosine similarity \(c_d\) is calculated for batch items with the same score but different grade level and batch items with the same grade level but different score as follows.

\begin{equation} \label{eq:10}
\begin{aligned}
c_d = {\sum_{g_i \neq g_j \in G}^{}\sum_{\substack{s_i=s_j \in S, \\ i \neq j}}Cos(z_i, z_j)} \\
+ {\sum_{\substack{g_i = g_j \in G, \\ i \neq j}}^{}\sum_{s_i \neq s_j \in S}Cos(z_i, z_j)}
\end{aligned}
\end{equation}
Finally, Prompt 5-6 model is trained to minimize the following loss function which incorporates both the MSE from Equation \ref{eq:8} and cosine similarities computed in Equations \ref{eq:9} and \ref{eq:10}.

\begin{equation} \label{eq:11}
L = MSE - \frac{1}{k}(c_s - c_d)
\end{equation}

The contribution of Grade Match is measured with test set QWK, and the results are compared against baseline models trained without Grade Match.

\subsubsection{Response Distortion} A service ready AES model must be able to detect and penalize incoherent word ordering. Our experiment investigates whether supervised training on the ASAP dataset alone leads to such results. Moreover, we experiment with a data augmentation method, Response Distortion, for adversarial training. Since adversarial input detection is applicable to all scoring rubrics, Response Distortion is applied to all six AES models.

Response Distortion generates a partially permuted essay sample from a normal essay response. Compared to similar works in AES adversarial training, Response Distortion is unique in that it only augments essays with a particular label score. For a given training batch \(b=\{t_1, t_2, ..., t_n\}\), we first filter essay samples with the lowest label score to get \(b'=\{t'_1, t'_2, ..., t'_m\}\) where \(t'_i=\{essay, prompt, score=0\}\). Next, we randomly select maximum of \(k\) samples from the filtered set \(b'\) for Response Distortion. For each selected \(k\) sample, we count the number of words \(w\) in the essay and select two indices \(i\) and \(j\) such that \(0 \le i < j \le w\). Finally, we randomly permute the ordering of all words between the \(i\)th and \(j\)th word index of the essay and add the generated distorted sample to the training batch with known label of score zero (lowest possible score). Since essays from \(b'\) are already assigned with the lowest label score, any distortions that further lowers the quality of writing will not change the assigned label score. Following the same logic, Response Distortion is also applied to prompt mismatched samples generated from Prompt Swap to introduce to the AES model various types of traits shared by low quality essay responses.

The contribution of Response Distortion is reported in two folds. First, \textbf{distorted response detection rate} is measured with distorted samples generated from the test set. Unlike the training stage in which Response Distortion is only applied to essays with the lowest label score, distortion is applied to essays without score condition during testing. Moreover, since Response Distortion only applies partial permutation to word ordering, we cannot assign the lowest label score to distorted test samples with certainty. Therefore, given normal essay \(t\), distorted essay \(t'\), and AES model \(f(essay) \rightarrow score\), distorted response detection is successful when \(f(t) > f(t')\) \cite{kabra2022evaluation}. Distorted response detection rate is computed and compared against baseline models trained without Response Distortion. Second, in response to previous work that reported a trade-off between anomaly detection rate and QWK performance \cite{ding-etal-2020-dont}, we test how Response Distortion affects test set QWK and compare the results against baseline models trained without Response Distortion.
\section{Results and Analysis} \label{results-and-analysis}
In this section, we analyze and discuss our experiment findings. Specifically, we evaluate the performance metric of each data augmentation method against baseline performances and examine the contribution of each method in terms of AES application. Moreover, we test a Response-Prompt AES model trained with our proposed data augmentations on the ASAP dataset and investigate how the results compare against those of previous neural approaches. Statistical significance is computed using paired t-tests between augmentation and baseline results. Statistical significance is denoted by $\boldsymbol{\cdot}$ for \(p < 0.1\), * for \(p < 0.05\) and ** for \(p < 0.01\).

\subsection{Data Augmentation Results}
\subsubsection{Prompt Swap}
Experiment result for Prompt 3-4 and Prompt 5-6 models trained with Prompt Swap is compared against two baseline models using irrelevant response detection rate as performance metric. The first baseline implements the Response-Prompt AES model structure and includes a response-prompt attention layer. However, the first baseline model is trained without Prompt Swap. The second baseline is a replicated model with the same model structure implemented in previous research, which consists of a regression layer attached to a pre-trained language model. Consistent with previous works, the second baseline model does not use the question prompt as input and is trained without Prompt Swap. Irrelevant response detection test results are summarized in Table \ref{table:1}.

\begin{table}[h!]
\centering
\footnotesize
\begin{tabular}{l | c | c c}
\hline
 & Response & \multicolumn{2}{c}{Irrelevant Response} \\
 & Prompt & \multicolumn{2}{c}{Detection Rate} \\ \cline{3-4}
Baseline & Attention & 3-4 & 5-6 \\ 
\hline
w/o Prompt Swap & No & 2.5\% & 0.0\% \\
w/o Prompt Swap & Yes & 2.5\% & 0.0\% \\
\hline
w/ Prompt Swap & Yes & \textbf{100\%}** & \textbf{100\%}** \\
\hline
\end{tabular}
\caption{Mean test set irrelevant response detection rate reported in averaged percentages over 5 folds.}
\label{table:1}
\end{table}
Experiment results indicate both baseline models trained without Prompt Swap fail to detect irrelevant responses. In other words, baseline models predict non-zero points to completely irrelevant essays. In contrast, Response-Prompt AES model trained with Prompt Swap records perfect detection rate in the test set. The results also demonstrate that the response-prompt attention layer is only relevant when implemented with Prompt Swap.

Furthermore, we analyze the response-prompt attention scores computed during Prompt Swap to investigate what is being learned by the AES model. Our findings summarized in Table \ref{Appendix C.1} of Appendix \ref{appendix:appendix_C} show the learned attention mechanism closely resembles the decision making process of a human rater.

\subsubsection{Grade Match} Experiment result for Prompt 5-6 model trained with Grade Match is compared against baseline model using QWK as performance metric. The baseline model implements the Response-Prompt AES model structure but is trained without Grade Match. Grade Match test results are summarized in Table \ref{table:2}.
\begin{table}[h]
\centering
\small
\begin{tabular}{l | c c} 
\hline
 & \multicolumn{2}{c}{QWK} \\ \cline{2-3}
 & Prompt 5 & Prompt 6 \\ 
Baseline & (Grade 8) & (Grade 10) \\ 
\hline
w/o Grade Match & 0.818 & 0.823 \\
\hline
w/ Grade Match & \textbf{0.829}$\boldsymbol{\cdot}$ & \textbf{0.837}$\boldsymbol{\cdot}$ \\
\hline
\end{tabular}
\caption{Mean test set QWK for Prompt 5 (Grade 8) and Prompt 6 (Grade 10) computed over 5 folds. Performance levels are computed from test set segmented by grade level for better comparison.}
\label{table:2}
\end{table}

When an AES model trains from essay responses written by students from different grade levels, our experiment results indicate applying Grade Match leads to QWK improvements in both grade levels. Moreover, our results confirm student grade level is indeed a factor to be considered during rubric-specific model training.

\subsubsection{Response Distortion} Experiment result for Prompt 1, 2, 3-4, and 5-6 models trained with Response Distortion is compared against baseline models using distorted response detection rate as performance metric (Response Distortion is not applicable in Prompt 7 and 8 models due to lack of lowest label score data in each rubric-segmented dataset). All baseline models implement the Response-Prompt AES model structure but are trained without Response Distortion. Test results summarized in Table \ref{table:3} clearly indicate AES models trained with Response Distortion record higher detection rates than their baseline counterparts for both partial and whole permutations.

Test results from Prompt 1 model show the contribution of Response Distortion is relatively small in datasets with smaller portion of lowest label score samples (\textit{See}, Table \ref{Appendix A.1} of Appendix \ref{appendix:appendix_A}). However, test results from Prompt 3-4 and 5-6 models suggest having more samples for augmentation does not guarantee linear increase in Response Distortion contribution. Lastly, Prompt 2 model shows even when the vanilla model performs well against full permutation, Response Distortion is still effective when processing partial permutations.

\begin{table}[h!]
\centering
\scriptsize
\begin{tabular}{l | c | c c c c} 
\hline
 & Distort & \multicolumn{4}{c}{Response Distortion Detection Rate} \\ \cline{3-6}
Baseline & Rate & 1 & 2 & 3-4 & 5-6 \\
\hline
w/o R.D. & 25\% & 38.5\% & 42.8\% & 21.8\% & 10.3\%\\
\hline
w/ R.D. & 25\% & \textbf{41.6\%} & \textbf{44.8\%} & \textbf{43.0\%}* & \textbf{24.7\%}* \\
\hline
\hline
w/o R.D. & 50\% & 39.8\% & 65.2\% & 32.2\% & 17.7\%\\
\hline
w/ R.D. & 50\% & \textbf{43.9\%} & \textbf{74.2\%} & \textbf{75.6\%}** & \textbf{51.6\%}* \\
\hline
\hline
w/o R.D. & 100\% & 60.1\% & 100.0\% & 51.6\% & 26.8\% \\
\hline
w/ R.D. & 100\% & \textbf{65.3\%} & \textbf{100.0\%} & \textbf{97.5\%}** & \textbf{83.8\%}** \\
\hline
\end{tabular}
\caption{Mean test set distorted response detection rate reported in averaged percentages over 5 folds. Distort Rate is set during testing to control the level of permutation. For example, when Distort Rate is 50\%, permutation indices \(i\) and \(j\) are sampled to cover 50\% of the original response.}
\label{table:3}
\end{table}

\subsection{ASAP Performance} So far, experiment results indicate a Response-Prompt AES model trained with our proposed augmentation is equipped with functions necessary to handle rubric items overlooked by previous neural approaches to ASAP. Nonetheless, since QWK is still a key component of AES assessment, we investigate the relationship between the added functions and the AES model's performance on the ASAP dataset.

We evaluate six AES models corresponding to six unique rubrics provided in ASAP and compare the results against benchmark performances in Table \ref{table:4}. Evaluation result reveals the following: a Response-Prompt AES model trained with Prompt Swap, Grade Match, and Response Distortion record the highest average QWK on the ASAP dataset when compared to previous neural based and state-of-the-art approaches (\(0.797 > 0.794\)). Performance comparison analysis at the model level exhibits the following strengths and areas for improvements of our proposed method.

\begin{table*}
\centering
\scriptsize
\begin{tabular}{l|l|cccccccc|c}
\hline
\multicolumn{1}{c|}{} & \multicolumn{1}{c}{} & \multicolumn{8}{|c|}{ASAP Prompt ID} & \multicolumn{1}{c}{} \\ \cline{3-10}
Row & AES Model & 1 & 2 & 3 & 4 & 5 & 6 & 7 & 8 & Average \\ 
\hline
1 & \citet{taghipour-ng-2016-neural} & 0.821 & 0.688 & 0.694 & 0.805 & 0.807 & 0.819 & 0.808 & 0.644 & 0.761 \\
2 & \citet{dong-etal-2017-attention} & 0.822 & 0.682 & 0.672 & 0.814 & 0.803 & 0.811 & 0.801 & 0.705 & 0.764 \\
3 & \citet{yang-etal-2020-enhancing} & 0.817 & \textbf{0.719} & 0.698 & \textbf{0.845} & \textbf{0.841} & \textbf{0.847} & 0.839 & 0.744 & 0.794 \\
4 & \citet{muangkammuen-fukumoto-2020-multi} & \textbf{0.833} & 0.685 & 0.690 & 0.795 & 0.812 & 0.816 & 0.798 & 0.673 & 0.763 \\
5 & \citet{mathias-etal-2020-happy} & \textbf{0.833} & 0.681 & 0.698 & 0.818 & 0.815 & 0.821 & 0.806 & 0.699 & 0.771 \\
6 & \citet{jeon-strube-2021-countering} & 0.828 & 0.706 & 0.694 & 0.827 & 0.806 & 0.820 & 0.838 & \textbf{0.769} & 0.786 \\
\hline
\hline
7 & Response-Prompt AES & 0.823 & 0.707 & 0.695 & 0.816 & 0.818 & 0.823 & \textbf{0.842} & \textbf{0.763} & 0.786 \\
\hline
8 & \hspace{1.5mm}Response Distortion & \textbf{0.830}$\boldsymbol{\cdot}$ & \textbf{0.719}* & 0.699 & 0.821$\boldsymbol{\cdot}$ & 0.823 & 0.827 & - & - & - \\
9 & \hspace{1.5mm}Prompt Swap & - & - & 0.702 & 0.830* & 0.823 & 0.829 & - & - & - \\
10 & \hspace{1.5mm}Prompt Swap + Response Distortion & - & - & \textbf{0.716}** & \textbf{0.832}** & 0.824 & 0.834$\boldsymbol{\cdot}$ & - & - & - \\
11 & \hspace{1.5mm}Grade Match & - & - & - & - & 0.829$\boldsymbol{\cdot}$ & 0.837$\boldsymbol{\cdot}$ & - & - & - \\
12 & \hspace{1.5mm}Prompt Swap + Response Distortion + Grade Match & - & - & - & - & \textbf{0.833}* & \textbf{0.839}$\boldsymbol{\cdot}$ & - & - & - \\
\hline
13 & Response-Prompt AES + Best Augmentations & 0.830 & \textbf{0.719} & \textbf{0.716} & 0.832 & 0.833 & 0.839 & \textbf{0.842} & 0.763 & \textbf{0.797}\\
\hline
\end{tabular}
\caption{Test set QWK performance for Response-Prompt AES model trained without augmentation (row 7), Response-Prompt AES model trained with various combinations of augmentations (rows 8-13), and AES models proposed in related works (rows 1-6). Augmented samples are only utilized during training and not included in test set QWK computation.}
\label{table:4}
\end{table*}

\paragraph{Prompt 1 \& Prompt 2 Models} Rows 7 and 8 of Table \ref{table:4} indicate adding Response Distortion results in QWK performance gain in both prompts 1 and 2. Such finding goes against previous studies reporting a performance trade-off \cite{ding-etal-2020-dont} between distorted response detection rate and QWK. As described in the experiment procedures, Response Distortion is different from related works in that it only augments essays with the lowest label score during training. The score condition is essential as it resolves the ambiguous task of assigning a score label to the generated sample without compromising overall label consistency. In line with related works, we confirm removing the score condition and extending Response Distortion to all score labels results in QWK performance loss.

\paragraph{Prompt 7 \& Prompt 8 Models} Response Distortion cannot be applied to datasets with insufficient number of low-score samples. Therefore, Prompt 7 and 8 model training is conducted without augmentations. Without augmentations, QWK performance in prompts 7 and 8 can be attributed to the Response-Prompt AES model structure as shown in Row 7 of Table \ref{table:4}. Scoring rubric for prompt 7 deducts points based on the essay's focus on the topic.\footnote{\url{https://www.kaggle.com/competitions/asap-aes/data}} Compared to benchmark models that only utilize the essay as input, Response-Prompt AES model utilizes both the essay and prompt as inputs and measures the essay's congruence with the prompt via response-prompt attention. Prompt 8 includes long essays that cannot be processed by previous approaches that inherit the 512 token length restriction from BERT. Response-Prompt AES model trains a self-attention layer without input length restriction to process longer essays and achieve better performance. However, \citet{jeon-strube-2021-countering} suggests adopting a pre-trained language model without length restriction can also be an alternative to training a custom layer from scratch.

\paragraph{Prompt 3-4 Model} While \(18\%\) of essays written in response to prompt 4 have zero label scores, only \(2\%\) of essays in prompt 3 have zero label scores, which makes low-score predictions particularly difficult in prompt 3. However, Prompt 3-4 model (i.e., rubric-specific model) is resilient to the label imbalance problem as it has access to zero label data from both prompts 3 and 4. Therefore, consistent with our findings from distorted response detection, we expect having access to sufficient number of zero label data will be an advantage for Prompt 3-4 model during augmentation training. Rows 8, 9, and 10 of Table \ref{table:4} not only demonstrate the individual effect of each augmentation, but also show Prompt Swap complements Response Distortion by generating additional low score samples for distortion, resulting in QWK improvement especially in prompt 3.

\paragraph{Prompt 5-6 Model} Rows 7 and 11 of Table \ref{table:4} confirm our hypothesis regarding rater expectation of writing quality and student grade level. Despite the QWK performance gain from Grade Matching, Prompt 5-6 model is outperformed by \citet{yang-etal-2020-enhancing} in both prompts 5 and 6. The results are aligned with the idea that prompt-specific models, when compared to generic models, are optimized to be the better performing model for a given prompt \cite{chen2013automated}. However, our results also indicate in exchange for prompt-specific performance, rubric-specific models benefit from efficiency \cite{Attali_Burstein_2006} as summarized in Table \ref{Appendix C.2} of Appendix \ref{appendix:appendix_C}.
\section{Conclusion}

In this paper, we seek to resolve the limitations of prompt-specific models while maintaining notable performance in the ASAP dataset. As a solution, we propose rubric-specific model training, which consists of a custom designed AES model trained from rubric-segmented datasets with series of data augmentations called Prompt Swap, Grade Match, and Response Distortion. Finally, we show the resulting AES model is capable of irrelevant response detection, student grade level adjustment, and distorted response detection while achieving state-of-the art performance in the ASAP dataset.

\section{Limitation}
Throughout this research, we identified several limitations relating to the performance metric and dataset utilized in the experiment.

First, while our research evaluates statistical significance among internal experiment results, statistical significance test was not applicable against external benchmark performances due to unavailability in released source code and limitations in replication.

Second, the ability to detect irrelevant response is a fundamental and expected feature of AES systems. Nevertheless, our experiments have demonstrated that the ASAP dataset and QWK do not test such fundamental attributes of AES models. Moreover, QWK provides limited information regarding the AES model's performance in other expected features such as fact checking or negation detection. Accordingly, while our experiment attains notable QWK performance in the ASAP dataset, we have insufficient understanding of our AES model's expected behavior against various data augmentation methods likely to be observed during real-world application.

Lastly, the QWK performance metric may not be aligned with the purpose of real-world application of autograding systems. Given that student performance in any academic field is mostly populated around the average, accurate evaluation is essential to identify the relatively smaller population of students who are falling behind or displaying talent. QWK is not an ideal performance metric for this purpose as capturing the majority around the average is a better strategy than capturing the small groups at both ends of the performance grid.

The ASAP dataset is one of many problems needed to be solved before real-world application of autograding systems. The limitations described above will be further discussed in our future work on AES focusing on service applications.

\bibliography{anthology}
\bibliographystyle{acl_natbib}

\appendix
\clearpage
\section{Data Tables}
\label{appendix:appendix_A}

\begin{table}[h!]
\small
\centering
\begin{tabular}{c c c c c c c c} 
 \hline
   & No. & Score & Avg. & Lowest Label \\
 Prompt & Essays & Range & Length & Score \% \\ 
 \hline
 1 & 1,783 & 2-12 & 350 & 0.56\% \\ 
 2 & 1,800 & 1-6 & 350 & 1.33\% \\
 3 & 1,726 & 0-3 & 150 & 2.26\% \\
 4 & 1,772 & 0-3 & 150 & 17.61\% \\
 5 & 1,805 & 0-4 & 150 & 1.33\% \\
 6 & 1,800 & 0-4 & 150 & 2.44\% \\
 7 & 1,569 & 0-30 & 250 & 0.0\% \\
 8 & 723 & 0-60 & 650 & 0.0\% \\
\hline
\end{tabular}
\caption{Summary statistics of the ASAP dataset. Score Range column indicates integer range of score labels. Lowest Label Score Percentage measures the portion of essays assigned with the lowest label score for each prompt. For example, in prompt 1, 0.56\% of 1,783 essays are assigned with the lowest label score of 2.}
\label{Appendix A.1}
\end{table}

\begin{table}[h!]
\centering
\begin{tabular}{c c c c c c c c} 
 \hline
 Prompt & Genre & Level & Rubric \\ 
 \hline
 1 & ARG & 8 &  \\ 
 2 & ARG & 10 & \\
 3 & RES & 10 & $\times$ \\
 4 & RES & 10 & $\times$ \\
 5 & RES & 8 & $\triangle$ \\
 6 & RES & 10 & $\triangle$ \\
 7 & NAR & 7 & & \\
 8 & NAR & 10 & & \\
\hline
\end{tabular}
\caption{Metadata of the ASAP dataset. Genre column indicates the type of essays including argumentative essays, response essays (source-dependent), and narrative essays. Level column indicates the grade level of the essay writers. Rubric column indicates prompts sharing the same scoring rubric. Scoring rubrics are identical for prompts 3 and 4 and prompts 5 and 6.}
\label{Appendix A.2}
\end{table}

\begin{table}[h!]
\centering
\begin{tabular}{c c c c} 
 \hline
 Prompt & Scoring Guide for Irrelevant Essay \\ 
 \hline
 3 & assign lowest score \\
 4 & assign lowest score \\
 5 & assign lowest score \\
 6 & assign lowest score \\
\hline
\end{tabular}
\caption{Scoring rubric for source dependant essays require evaluation of relevance between essay response and question prompt.}
\label{Appendix A.3}
\end{table}
\clearpage
\section{Hyper-parameters}
\label{appendix:appendix_B}

Our experiments are conducted with 4 NVIDIA GeForce RTX 3090 GPUs, and training batch size for each AES model is set to match the maximum GPU memory limit.

\paragraph{Prompt 3-4 Model} We train Prompt 3-4 model on the rubric-segmented dataset for 20 epochs. We apply learning rate of \(4 \times 10^{-5}\) for the pre-trained BERT and \(8 \times 10^{-5}\) for the custom attention layers which are trained from scratch. For accurate performance comparison, develop and test set performances are recorded and reported separately for each prompt. Training batch of size 40 is applied with \(5\%\) Prompt Swap rate, resulting in total of 42 training data samples for each batch.

\paragraph{Prompt 5-6 Model} Prompt 5-6 model is trained for 40 epochs in total, and cosine similarity is optimized for the first 20 epochs only. After 20 epochs, Prompt 5-6 model training only optimizes the MSE loss. To make sure a given training batch is sufficiently diverse, each batch item is paired with a positive and negative sample randomly selected from outside the training batch, resulting in training batch of size 16. We apply learning rate of \(8 \times 10^{-5}\) for the pre-trained BERT, \(1 \times 10^{-4}\) for the custom attention layers, and \(3 \times 10^{-4}\) for cosine similarity optimization. Training batch of size 16 is applied with 2 Prompt Swap samples per batch, but prompt swapped samples are excluded from cosine similarity computation.

\paragraph{Prompt 1, 2, 7, and 8 Models} Prompts 1, 2, 7, and 8 have distinct scoring rubrics and therefore are trained separately with distinct hyper-parameter settings. Response Distortion is applied to essay responses that have the lowest label scores when applicable. Learning rates ranging from \(1 \times 10^{-5}\) to \(4 \times 10^{-5}\) are applied for the pre-trained BERT and \(8 \times 10^{-5}\) to \(1 \times 10^{-4}\) for the custom attention layers. Batch size ranges from 10 to 16 with a response distortion rate of 1 augmented sample per batch over 10 to 20 training epochs.

\paragraph{Irrelevant Response Detection} During irrelevant response detection testing for Prompt 3-4 and Prompt 5-6 models, we apply Prompt Swap rate of \(10\%\) to generate 72 prompt mismatched test samples for each prompt and each fold. To better capture the contribution of Prompt Swap, Prompt Swap is only applied to essays with scores greater than the average score during testing. Prompt mismatched samples are only used to compute irrelevant response detection rate and are not included in test set QWK calculation.

\paragraph{Distorted Response Detection} During distorted response detection testing for Prompt 1, 2, 3-4 and 5-6 models, we apply Response Distortion to all samples in each rubric-segmented test set for all folds. Response Distortion is applied to essay samples without score conditions during testing. Moreover, the same test is conducted with different Distort Rate values. Distort Rates are only applied during testing to control the magnitude of permutation applied on each test sample. Distorted samples are only used to compute distorted response detection rate and are not included in test set QWK calculation.
\clearpage
\onecolumn
\section{Additional Results}
\label{appendix:appendix_C}

\begin{table}[h]
\centering
\begin{tabular}{c | c}
\hline
Attention Score & Sentences from Question Prompt \\
\hline
0.0368 & ``Winter Hibiscus by Minfong Ho Saeng, a teenage girl, and her family \\
& have moved to the United States from Vietnam.'' \\ [1ex] 
0.0325 & ``A wave of loss so deep and strong that it stung Saeng's eyes now swept \\
& over her.'' \\ [1ex] 
0.0280 & ``I'd read once that sucking on stones helps take your mind off thirst by \\
& allowing what spit you have left to circulate.'' \\
\hline
\hline
0.0019 & ``Write a response that explains why the author concludes the story \\
& with this paragraph.''  \\ [1ex] 
0.0029 & ``How did it go? Did you-?''  \\ [1ex] 
0.0029 & ``Goodness, it's past five. What took you so long?'' \\
\hline
\end{tabular}
\caption{Response-prompt attention scores computed during Prompt 3-4 model training with Prompt Swap. Table includes three largest and three smallest attention score values with their corresponding question prompt sentence. Sentences corresponding to the first two largest attention scores can be easily associated with prompt 4, which is a story describing the struggles of immigration. Similarly, the third largest attention score can be associated with prompt 3, which is an essay describing a cyclist's battle against thirst and dehydration. On the contrary, prompt sentences with the lowest attention scores cannot be directly associated with either prompt 3 or prompt 4.}
\label{Appendix C.1}
\end{table}

\begin{table}[h]
\centering
\begin{tabular}{l | c c | c}
\hline
 & \multicolumn{2}{|c|}{QWK} & \\ \cline{2-3}
AES Model & 3-4 & 5-6 & No. of Trained Models \\
\hline
\citet{yang-etal-2020-enhancing} & 0.772 & 0.844 & 4 \\
\citet{jeon-strube-2021-countering} & 0.761 & 0.813 & 4 \\
\hline
\small{Rubric-Specific Model w/ Augmentation Training} & 0.774 & 0.836 & \textbf{2} \\
\hline
\end{tabular}
\caption{Test set QWK performance (prompt averages) and number of trained AES models for benchmark models utilizing pre-trained language models. Prompt-specific approach requires training \(n\) models for \(n\) question prompts. On the other hand, rubric-specific approach only requires training 1 model for \(n\) question prompts as long as the prompts share the same rubric.}
\label{Appendix C.2}
\end{table}

\end{document}